# An Overview of Melanoma Detection in Dermoscopy Images Using Image Processing and Machine Learning

Nabin K. Mishra[a] and M. Emre Celebi[b]

[a] *Stoecker & Associates, Rolla, MO, USA*
[b] *Department of Computer Science, Louisiana State University, Shreveport, LA, USA*

January 27, 2016

## Abstract

The incidence of malignant melanoma continues to increase worldwide. This cancer can strike at any age; it is one of the leading causes of loss of life in young persons. Since this cancer is visible on the skin, it is potentially detectable at a very early stage when it is curable. New developments have converged to make fully automatic early melanoma detection a real possibility. First, the advent of dermoscopy has enabled a dramatic boost in clinical diagnostic ability to the point that melanoma can be detected in the clinic at the very earliest stages. The global adoption of this technology has allowed accumulation of large collections of dermoscopy images of melanomas and benign lesions validated by histopathology. The development of advanced technologies in the areas of image processing and machine learning have given us the ability to allow distinction of malignant melanoma from the many benign mimics that require no biopsy. These new technologies should allow not only earlier detection of melanoma, but also reduction of the large number of needless and costly biopsy procedures. Although some of the new systems reported for these technologies have shown promise in preliminary trials, widespread implementation must await further technical progress in accuracy and reproducibility. In this paper, we provide an overview of computerized detection of melanoma in dermoscopy images. First, we discuss the various aspects of lesion segmentation. Then, we provide a brief overview of clinical feature segmentation. Finally, we discuss the classification stage where machine learning algorithms are applied to the attributes generated from the segmented features to predict the existence of melanoma.

## Introduction

Malignant melanoma, referred to herein as "melanoma," is a type of cancer that in almost all cases starts in pigment cells (melanocytes) in the skin. In this review, we omit relatively rare primary melanomas beginning outside the skin; in the brain, eye, or mouth. Most persons when they think of melanoma think of a dark raised lesion; after all these tumors grow from pigment cells. But some melanomas lose pigment, lacking the dark pigment partially or completely, and appear pink, white or tan [1]. Another departure of many recent early melanomas from the stereotypical raised dark prototype of the past is that they are often flat. Apart from coloration and elevation, there are many other characteristics of melanomas such as texture and existence of certain structures (clinical features) in the lesion that differentiates them from benign lesions.

Melanoma is considered to be the deadliest of skin cancers. Although for a given case Merkel cell carcinoma is more likely to be fatal, melanoma overall causes more deaths than any other type of skin cancer [2]. According to the American Cancer Society (ACS), about 76,380 (46,870 men and 29,510 women) new cases of melanomas are estimated to be diagnosed in the year 2016 [3]. For the same year, ACS also estimated about 10,130 fatalities (6,750 men and 3,380 women) [3]. The incidence of melanoma has been rising every year. Many of these lives could be saved if melanoma were to be detected at the earliest stage, when it is easily curable. A number of studies using various technologies are being conducted around the world for the early detection of melanoma.

Dermoscopy (also known as dermatoscopy or epiluminescence microscopy) is a method of acquiring a magnified and illuminated image of a region of skin for increased clarity of the spots on the skin [4]. The imaging instrument used for this purpose is called a dermatoscope. Dermatoscopes are of two types: contact, using a layer of gel/oil applied between skin and dermatoscope, and non-contact, with no skin contact and no fluid. Non-contact images, and some contact images, use cross-polarized light from the dermatoscope to acquire the image. Dermoscopy images, because of their illumination and magnification, are widely used in the analysis and examination of skin lesions.

There are very few publicly available datasets of dermoscopy images. Among them, PH$^2$ [5] and EDRA [6] image databases are most commonly used by the research communities. Recently, the International Skin Imaging Collaboration (ISIC) has also created the ISIC Archive for the Melanoma project which is a large public database of dermoscopy images [7]. The dermoscopy images from these databases can be used for the research, development and comparison of various algorithms for identifying melanoma. Figure 1 demonstrates an overall schematic of the steps involved in the identification of melanoma using dermoscopy images. Three major stages can be identified in this diagram: lesion segmentation, feature segmentation and classification [8] [9] [10].

## Lesion Segmentation

A skin lesion, in a dermoscopy image, is a single bounded region that is most often distinguishable from the normal surrounding skin by virtue of different color or texture; this area is considered to be the region of interest for further processing. Segmentation of the lesion means separating that region (lesion) from the normal skin region (non-lesion). Lesion segmentation is a very important step in the analysis of dermoscopy images for it allows the identification of various global morphological features specific to the lesion and at the same time provides a confined region for segmentation of various local clinical features at a later stage. The edge of the segmented region, called the border or boundary, also provides features for use in the analysis of the lesion [11]. Correct identification of the non-lesional area, ignoring artifacts present in some images, also provides a region of normal skin for calculating relative colors and other useful features [12] [13].

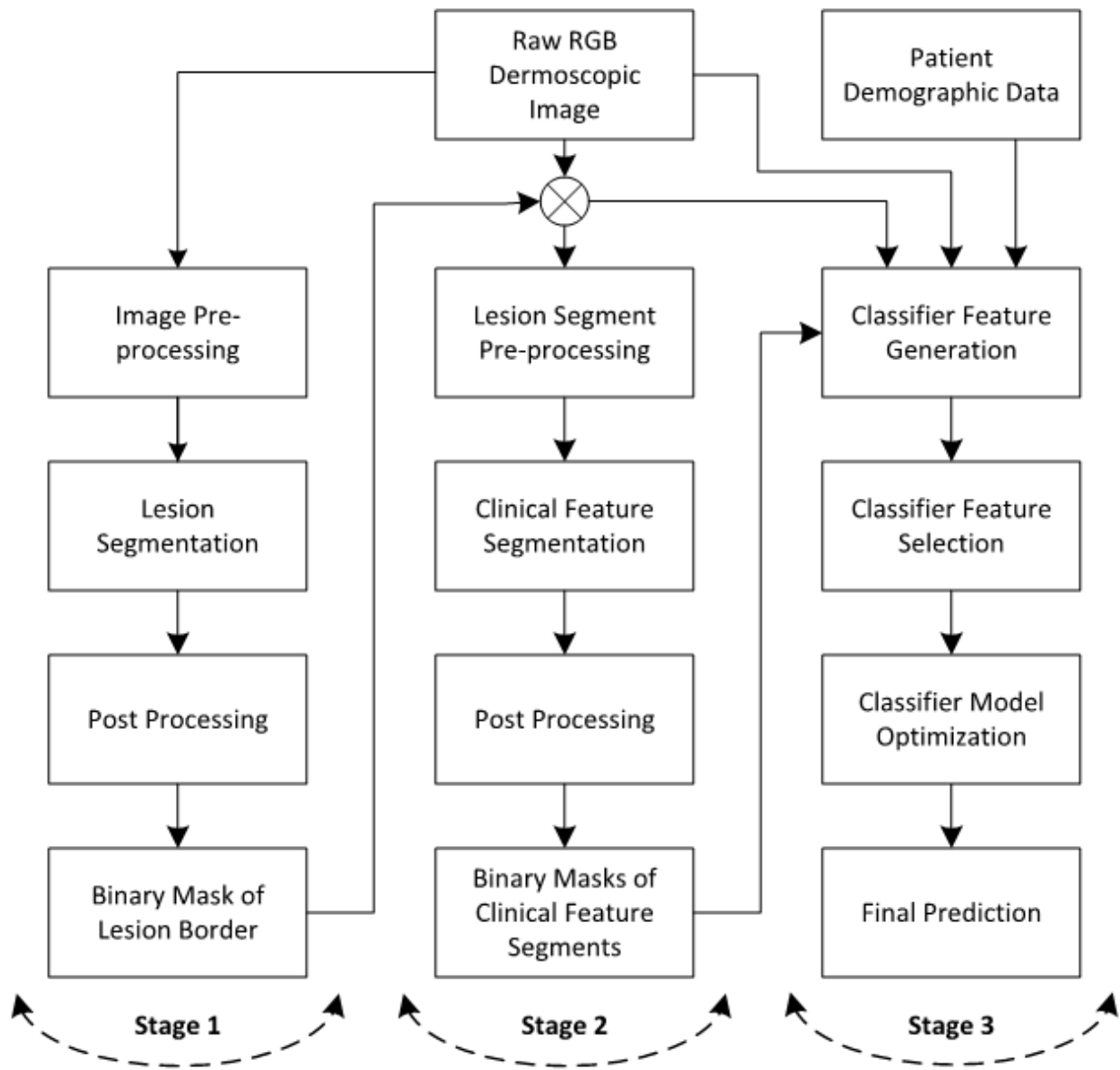

Figure 1: A general block diagram of computerized detection of melanoma in dermoscopy images. Many details are involved in each of the blocks shown.

Lesion segmentation is a very challenging problem for several reasons. The foremost reason is the relatively poor contrast between the normal and lesional skin in many cases. Other reasons include variations in skin tone, skin aberrations including presence of artifacts (hairs, ink, bubbles, ruler marks, date markers, color calibration charts, etc.), non-uniform lighting, non-uniform vignetting (exterior black circles), physical location of the lesion and most importantly variations in the lesion itself in terms of color, texture, shape, size and location in the image frame. Each of these factors should be considered while designing a robust lesion segmentation algorithm [14] [15]. The influence of most of these can be minimized by proper pre-processing steps in lesion segmentation.

Figure 2 shows various dermoscopy images with different artifacts and aberrations. Figure 3 shows additional images with variations in skin tones and in lesion characteristics.

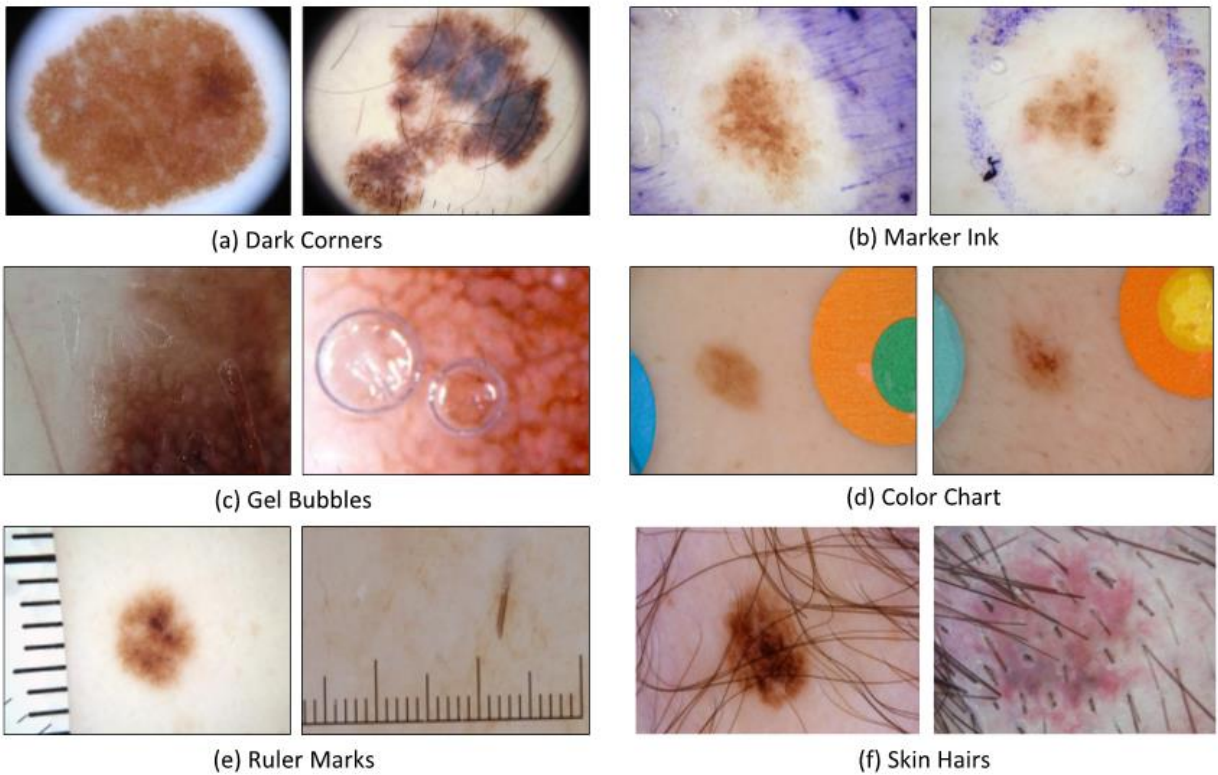

Figure 2: Various artifacts seen in dermoscopy images [7]

Basic pre-processing steps include eliminating variable lighting effects [16] [17] [18] [19], converting the image into a different color space [20], selecting an appropriate color channel [21], enhancement of the selected color channel [22], contrast enhancement [23] [24], normalizing color variation caused by image acquisition, smoothing the image, removing hairs [25] [26] [27] [28] [29], removing the vignetting effect [30], and localizing the lesion [31]. A proper combination of pre-processing steps can play a significant role in accurate lesion segmentation.

There are various broadly categorized image segmentation methods including but not limited to methods employing histogram thresholding [21] [32] [33], clustering [34] [35] [36] [37] [38], localized and distributed region identification, active contours (snakes) [39] [40], edge detection [41] [42], fuzzy logic, supervised learning, graph theory [43], and probabilistic modeling [44] [45]. These methods can be applied individually or in combination to achieve maximum accuracy [14] [46] [47].

Post-processing is essential for accurate lesion segmentation and feature detection and generation. Common post-processing methods include region merging [44] [45], smoothing, opening and closing operations, peninsula detection and removal [48], island removal, and border expansion.

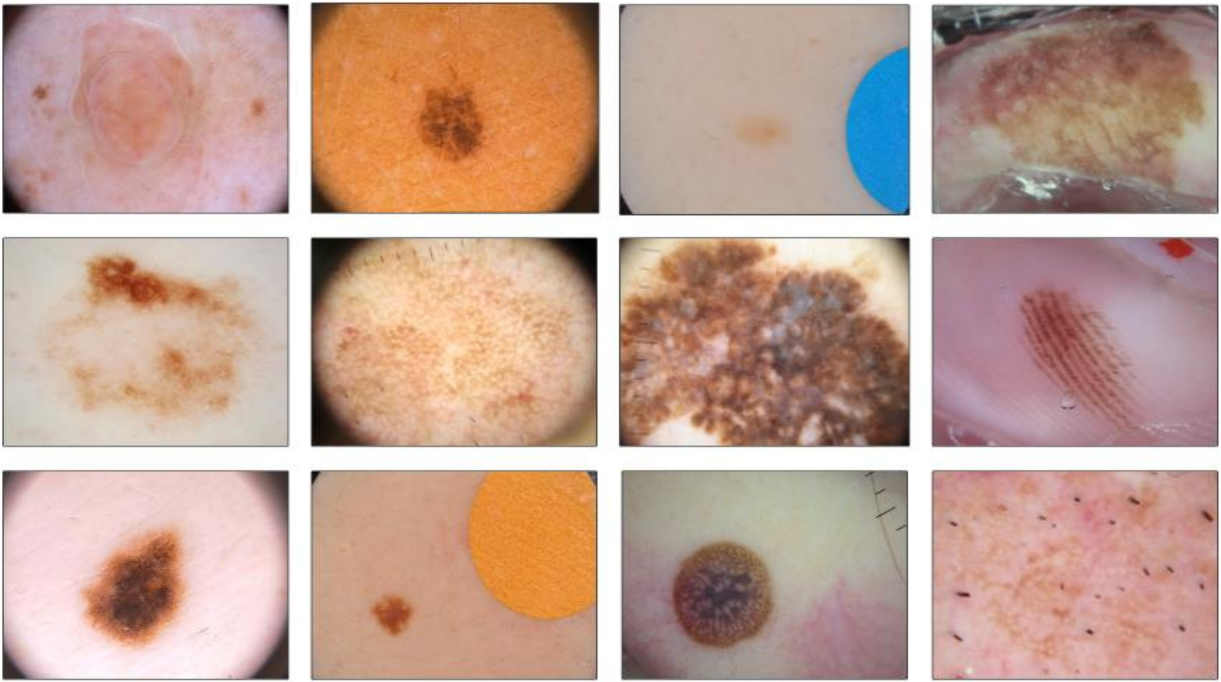

Figure 3: Dermoscopy images illustrating variations in lesion characteristics [7]

Evaluation of the results of lesion segmentation algorithm can be equally complex and debatable in terms of scoring the best lesion segmentation. Lesion segmentation varies widely among experts as well. In most cases, a gold-standard lesion segmentation is created using multiple lesion regions obtained from different human experts; the automated border is compared to a standard manual border obtained by combining individual manual borders. Evaluation methods can be subjective or objective. A subjective evaluation is based on a score provided by human subjects. In this case the experimental setup for scoring should be properly defined such that the scoring method is consistent among different subjects. An objective evaluation is based on error quantities calculated while comparing the automated border with the standard manual lesion boundary. Various objective evaluation measures are defined in the literature and each has its own advantages and disadvantages [49].

## Feature Segmentation

In a given dermoscopy image of a lesion, there may exist multiple clinical features that indicate whether the lesion is benign or malignant. A given feature may be global, spanning the lesion area; local, present in a small area; or present at multiple spots in the lesion. So, unlike lesion segmentation, each feature segmentation in most cases will have multiple segments around the lesion area. Pigment network [50], atypical and typical networks, streaks [51] [52] [53], regression structures [54], starburst pattern, dots and globules [55], blotches [56], blue-white veil [57] [58] [59], pink shades [60], white areas [61], milia-like cysts, vascular structures [62], etc. are some of the commonly employed features in predicting melanoma [63] [64] [65]. Most of these feature structures can also be broadly categorized as having different patterns such as reticular, globular, cobblestone, homogeneous, starburst, parallel, multicomponent, lacunar and unspecific [63] [66] [67] [68]. While using these features for identifying melanoma, benign dermoscopic features in melanomas can be equally important in automatic identification [69].

As in lesion segmentation, feature segmentation also includes pre-processing, segmentation and post processing steps. Color, texture, shape, structure, relative size, location in the lesion etc. are some of the main attributes used in clinical feature segmentation [70]. In addition to their presence, distribution of a feature in the lesion area provides further diagnostic information. The steps involved in clinical feature segmentation are very similar to the lesion segmentation steps described in the previous section.

Feature pre-processing steps are largely feature-dependent. Pre-processing techniques such as color standardization/correction and correcting lighting variation are similar to the equivalent techniques used in lesion segmentation. Any method involving image enhancement, sharpening, blurring, color space transformation, frequency/space transformation, etc., would be dependent on the type of feature being segmented. A number of such pre-processing methods are rigorously investigated upon through many iterations to find the right combination that best suits the detection of the targeted feature. Similar analysis is performed to select the best color, texture or frequency channel. Either visual observation or annotated masks can be used for tuning/training in these steps.

During the feature segmentation step, a different approach is taken to deal with the artifacts present in the image. Artifacts such as hairs and gels often occlude the targeted feature. Accordingly, hair or gel masks are used either in pre-processing (avoiding those regions for further analysis) or post-processing (using the mask to get rid of any feature segments found in these regions), depending on the segmentation approach chosen. These hair or gel masks can also be created automatically which in itself would be a separate segmentation problem. Ruler marks are dealt with similarly and they mimic hair-like structures, but are usually straight and shorter. Sometimes hairs are also white and thicker. Artifacts such as dark peripheral regions and color wheels should be avoided entirely in any analysis by masking them out in advance. This may already be achieved in the initial lesion segmentation step.

All types of segmentation algorithms mentioned above in the context of lesion segmentation can also be used for feature segmentation. However, the final feature segmentation output will have multiple distributed segments of different shapes and sizes based on the feature being segmented. Choosing the right combination of color channels for applying the segmentation algorithm is equally important as choosing the right segmentation approach. The post-processing in this case is also crucial and should be selected carefully based on the kind of filtering required to achieve best results. It should be kept in mind that features used in the ultimate classification of the lesion will be generated later using the feature segments.

The evaluation criteria are exactly the same as those discussed in the lesion segmentation section except that there may be a number of images where the target feature does not exist and the algorithm should not be finding anything. So, an extra step of evaluation can be performed by calculating the binary success/failure rate of feature segmentation.

## Feature Generation and Classification

Predicting a lesion to be benign or malignant is a binary classification problem. In order to solve a classification problem, features or attributes that characterize the samples are required. In the problem of melanoma detection, some features can be collected clinically and some are generated using dermoscopy images. Lesion and feature segmentation are the preliminary steps in the feature generation process.

Lesion-related morphological features such as estimated diameter (estimated because the images may be acquired at different magnification levels), symmetry, irregularity, eccentricity etc. can be calculated from the lesion border. Color and texture features related to the lesion can also be calculated from the lesion area; these can be referred to as global features. Various color channels can be used for this purpose. Because of the significance of various color distribution and texture around the lesion, clustering methods can also be used to divide the lesion into various regions and then color and texture features can be calculated from those regions separately. The lesion area may also be divided concentrically into various peripheral and central regions from which global features can be extracted.

In order to generate a clinical feature, specific attributes and/or masks created using feature segmentation are used. Color and texture features are very commonly used in this case as well. In addition, morphological attributes of such segments such as shape, size, location in the lesion etc. may also be used. In some cases, it is also important to examine the attributes of the surrounding regions for proper discrimination of melanoma.

The final step is to use these attributes (global and feature specific) in a classifier to be able to distinguish melanomas from benign lesions. A number of classifiers are available for experimentation and they can be chosen based on their performance; the final model can be tuned for better performance [71]. Some of the very commonly used classifiers are artificial neural networks (ANNs) [72] [73], support vector machines (SVMs) [74], logistic regression [75], decision trees [76], ensemble learners [77] [78], k-nearest neighbors (kNN) [79], Bayesian classifiers, deep learning algorithms [80], discriminant analysis [81], etc. There are numerous other classifiers that can also be explored for classification.

The evaluation of the classifier result is based on the overall accuracy, sensitivity and specificity of the system on the test set. In this domain, it is very important not to miss melanoma while being able to correctly identify benign lesions as much as possible. In other words, the ultimate goal is to target highest sensitivity while optimizing to increase specificity, thereby increasing the overall accuracy.

## Conclusion

The early detection of melanoma is essential for successful treatment. Because dermoscopy images are so inexpensive to obtain and so widely available, they provide the most viable option for application of new image processing and machine learning algorithms. Therefore, melanoma detection using dermoscopy images has the most potential for disruption of the current clinical paradigm of waiting until the melanoma is at a later stage and performing an excessive number of biopsies. The advent of a fast, accurate and cost-effective on-the-spot technology, in the clinic or even at home, is most likely to be afforded by the type of computer analysis of dermoscopy images described here. Dermoscopy images come with various aberrations and artifacts and hence it is crucial to follow the proper steps and methods described here to remedy these abnormalities and achieve a correct diagnosis. Lesion segmentation with acceptable tolerance allows for acceptable precision in feature segmentation which in turn helps in maximizing classification accuracy. Although lesion segmentation, feature segmentation, feature generation and classification are the major steps, proper attention should be given to the auxiliary steps which in most cases are the major contributors to an exemplary outcome.